# CNER: A tool Classifier of Named-Entity Relationships

POSTER

Jefferson A. Peña Torres *
Universidad del Valle
Cali, Colombia

Raúl E. Gutiérrez De Piñerez
Universidad del Valle
Cali, Colombia

**Abstract**: We introduce CNER, an ensemble of capable tools for extraction of semantic relationships between named entities in Spanish language. Built upon a container-based architecture, CNER integrates different Named entity recognition and relation extraction tools with a user-friendly interface that allows users to input free text or files effortlessly, facilitating streamlined analysis. Developed as a prototype version for the Natural Language Processing (NLP) Group at Universidad del Valle, CNER serves as a practical educational resource, illustrating how machine learning techniques can effectively tackle diverse NLP tasks in Spanish. Our preliminary results reveal the promising potential of CNER in advancing the understanding and development of NLP tools, particularly within Spanish-language contexts.

## 1. Introduction

In the Natural Language Processing (NLP) Group at Universidad del Valle context, English serves as the main language to learn the fundamentals of how Machine Learning (ML) techniques work to address various NLP tasks. However, Spanish is occasionally adopted as the focus language for research endeavors and as result multiple projects are conducted in Spanish to explore language-specific nuances and challenges in NLP applications. Named-Entity recognition [1], Machine Translation [2], Semantic Relation Extraction [3] among others tasks have been conducted with a focus on Spanish language data, allowing for a more nuanced understanding of the intricacies involved.

In this paper we present Classifier for Named Entities Recognized (CNER) a linguistically-aware online service that offers the possibility to test two main tasks of NLP, Named Entity Recognition (NER) and Relation Extraction (RE) for Spanish language. This together with other projects on Spanish language have been evaluated and adapted as a web service. In this context, language technologies and natural language processing (NLP) tools can support the identification of useful information in text and to promote its understanding. Specifically, CNER i) identifies the mentions follow the ACE standard with entity types include Person (PER), Organisation (ORG), Facility (FAC), Location (LOC), Geographical/Political (GPE), Vehicle (VEH), Vehicle (VEH) and Weapon (WEA) [4], [5]; ii) displays three different NER tools as previous step to RE task and iii) offers entity relationship information through tags GPE-AFF, PHYS, DISC, EMP-ORG, ART, NON-REL representing the relations between two entities [6].

* jefferson.amado.pena@correounivalle.edu.co
Corresponding author

This paper is structured as follows: In Section 2 we present the related work; in Section 3, we detail the design criteria, technical specifications and software principles to develop CNER and we explain its functionalities in Section 4. We discuss the scope of the application in Section 5 and we conclude and outline the future work in Section 6. We set an example as shown in Appendix A.

## 2. Related Work

In the last years, several open-source NLP tools have been published, being available to the users. Some of them provide models for languages such as English, Arabic, Portuguese and others varieties in Spanish, Chinese and German languages. There are several tools available such as Stanford CoreNLP [7] which is one suites that include tokenizers, PoS-taggers, named entity recognizers mainly for English and recently for Spanish, Chinese, German and Arabic, Some CoreNLP modules has been published as a online demos improve success and popularity[1] [2] [3].

Another popular tool is FreeLing [8], which includes similar modules than CoreNLP and several language analyzers written in C++ is focused for Catalan, Spanish and recently for Portuguese, English, among others. FreeLing has the a useful demo as web application[4] that allows testing several NLP analysis. In the same way, there are some toolkits that have been incorporated in a web service, web application or demos. NLTK [9] is a popular toolkit written in Python that includes sentiment analysis, part-of-speech tagging, phrase chunking, named entity recognition, text classification, stemming and tokenization and has been deploy to test some modules online[5]. Another toolkit is OpenNLP [10] written in Java which performs most common NLP tasks with some models for several languages, including English, Spanish or German.

CNER presented in this paper follows the software engineering principle in which the main components have an interface as overall architecture. It provides a simple, efficient and ready to use set of NLP tools in Spanish language.

## 3. CNER Web Application

CNER comprises two integral components: a RESTful backend that seamlessly integrates with other services, ensuring scalability and seamless connectivity with the frontend; and a user-friendly interface designed to receive inputs and display results efficiently. Both components adhere to modern web development practices, facilitating enhanced usability and performance. Below, we provide a detailed description of each component, highlighting their functionalities and contributions to the overall system.

---

[1] https://corenlp.run/
[2] http://nlp.stanford.edu:8080/ner/
[3] http://nlp.stanford.edu:8080/parser/
[4] https://nlp.lsi.upc.edu/freeling/node/1
[5] http://text-processing.com/demo/

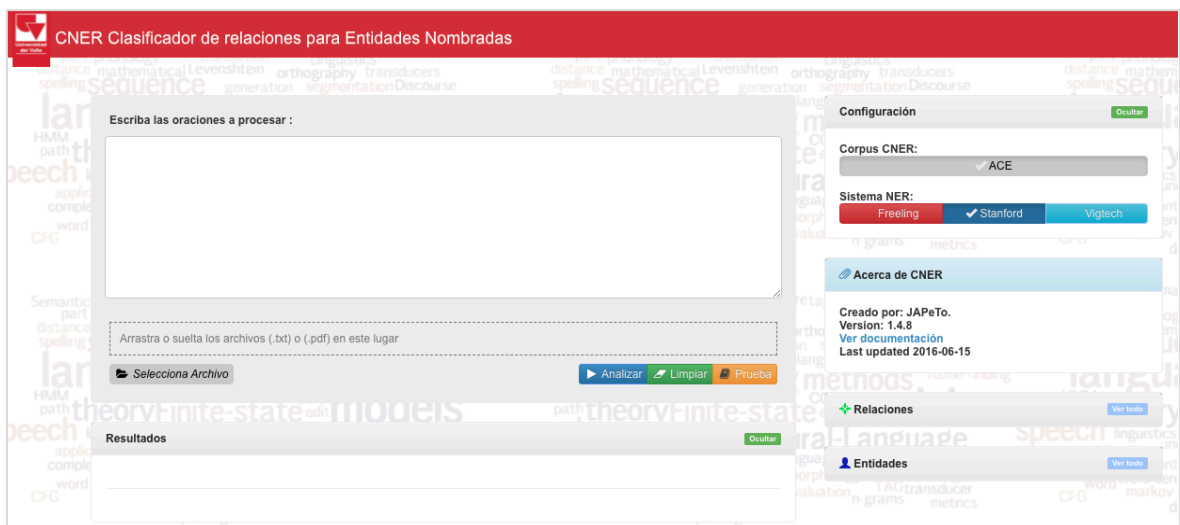

Figure 1: Main Page of CNER Web Tool. See http://eiscapp.univalle.edu.co/cner/

**Backend:** When choosing which environment of the back-end of the CNER prototype the choice was between hosting it on host or as a containerized application. The host was not chosen because it provides less freedom and requires specific service requirements. The back-end and several tools as Freeling, CoreNLP, Spanish NER [1], and SVM for Spanish [3] were organized as API REST services developed in Python language, while other Natural languages toolkits were used to process the inputs. An overall design is depicted in Figure 1.

Next, each stage is briefly explained.

- Input processing: Involves raw processing of electronic text with the help of NLP toolkits and includes tasks such as section splitting, tokenization, relationship extraction etc. At the end of this stage, output will contain the base representation from user input.
- Choosing the NER tools: In CNER the Named Entity Recognition (NER) could be taken by three available models for Spanish to find key elements such as persons, locations, organization as previous step to Relation Extraction.
- Relation extraction (RE): With the entity mentioned the extraction and classification of the important relationships among entities are returned as answer.

The scope of the CNER is to get the named entities relationships with a high precision and recall in any spanish text. If a user enters a new symptom then the system should be able to answer with the probable disease name.

**Front-end:** The user input interface, the front-end is a containerized application, that accepts query from the user. Electronic text can be a sentence from a news, a social network or any text in Spanish. The user interface will accept this text and give output as the entities and relationships. The technology used for web development is Flask Framework. Flask is a micro framework for Python based on Werkzeug, Jinja 2, that allows deployment of templates with HTML, CSS (Mainly, Bootstrap ) and JavaScript..With the help of flask part of input is processed.

The NER selection options are accessible and visible on the initial screen. The user only needs to set the corpus, NER tool and submit the text to get the results.

### 3.1. Functionalities

In this section we describe the CNER functionalities.
- Allows the user to insert text manually or upload files in any of the following formats: .txt, .doc or .odt. files.
- Allows texts in spanish languages. The user can select the named entity recognizer from three buttons on the initial screen.
- Computes and displays the mentions and its relationships according to the selected NER tool
- Offers an visual information sentence by sentences, the most important words and relationships
- Integrates textual external tools to assist in all process information and preprocessing raw text.
- Allow incorporation of new tools and models.

Its functionalities make the academic and research establishments as a service.

### 4. Evaluation and Results

As part of an NLP course at Universidad del Valle, we conducted a user evaluation to assess the effectiveness of CNER. Participants were required to analyze Spanish text to provide valuable feedback.

### 5. Conclusion and Future Work

The field of extracting mentions or entities from text using NLP is a heavily researched area. However, due to the unique characteristics of Spanish text, it requires specific strategies and tools as compared to English NLP tools. To address this need, we developed a prototype tool capable of extracting basic information about named entities as well as their relationships in Spanish text.
    The results of the evaluation showed that the entity recognition and its relationships as service is very useful. Although at the moment it only works for academic or research environments, it can be adapted to other domains, such as industrial or commercial. As future work, we want to adapt the tool to work with other available services in Spanish languages, as well as debugging possible errors on the classification and extraction.